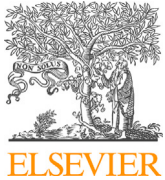
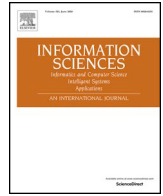

# Wearable-based behaviour interpolation for semi-supervised human activity recognition

Haoran Duan [a], Shidong Wang [b], Varun Ojha [c], Shizheng Wang [e,*], Yawen Huang [d], Yang Long [a,*], Rajiv Ranjan [c], Yefeng Zheng [d]

[a] *Department of Computer Science, Durham University, UK*
[b] *School of Engineering, Newcastle University, UK*
[c] *School of Computing, Newcastle University, UK*
[d] *Jarvis Research Center, Tencent YouTu Lab, China*
[e] *Chinese Academy of Sciences R&D Center for Internet of Things, China*

A B S T R A C T

While traditional feature engineering for Human Activity Recognition (HAR) involves a trial-and-error process, deep learning has emerged as a preferred method for high-level representations of sensor-based human activities. However, most deep learning-based HAR requires a large amount of labelled data and extracting HAR features from unlabelled data for effective deep learning training remains challenging. We, therefore, introduce a deep semi-supervised HAR approach, MixHAR, which concurrently uses labelled and unlabelled activities. Our MixHAR employs a linear interpolation mechanism to blend labelled and unlabelled activities while addressing both inter- and intra-activity variability. A unique challenge identified is the activity-intrusion problem during mixing, for which we propose a mixing calibration mechanism to mitigate it in the feature embedding space. Additionally, we rigorously explored and evaluated the five conventional/popular deep semi-supervised technologies on HAR, acting as the benchmark of deep semi-supervised HAR. Our results demonstrate that MixHAR significantly improves performance, underscoring the potential of deep semi-supervised techniques in HAR.

## 1. Introduction

It is well-known that body-worn sensors are used for many real-world applications, including but not limited to sleep monitoring [1], environmental perception [2], elderly patient assistance and health assessment [3], etc. Also, wearable-based human activity recognition (HAR) is one of the core research areas in ubiquitous computing, and it plays an essential role in human behaviour understanding, health monitoring, skill assessment, sports training, etc. [4]. In contrast to computer vision-based action recognition, the data is collected by using wearable sensors, making it less constrained and privacy-friendly.

Traditional feature engineering for HAR [3] tends to be a trial-and-error process, which may vary from task to task. Hence, deep learning became popular for high-level representations of sensor-based human activities [6,7]. However, most of the deep learning based HAR rely on the supervised learning paradigm, which requires substantial labelled data for model training. The unlabelled data in real-world scenarios can be collected easily, while the annotations are labour-intensive and time-consuming [8]. Methods that can learn behaviour patterns from unlabelled activities have drawn much attention in the HAR research community, e.g., self-supervised HAR [9], yet effectively learning diverse behaviour patterns from both unlabelled data may not be a trivial task. Fig. 1 shows the HAR model performance of different learning paradigms, and it seems that the feature learned by recent self-supervised work[1] [5] on the wearable-based dataset does not help much when compared with supervised training.

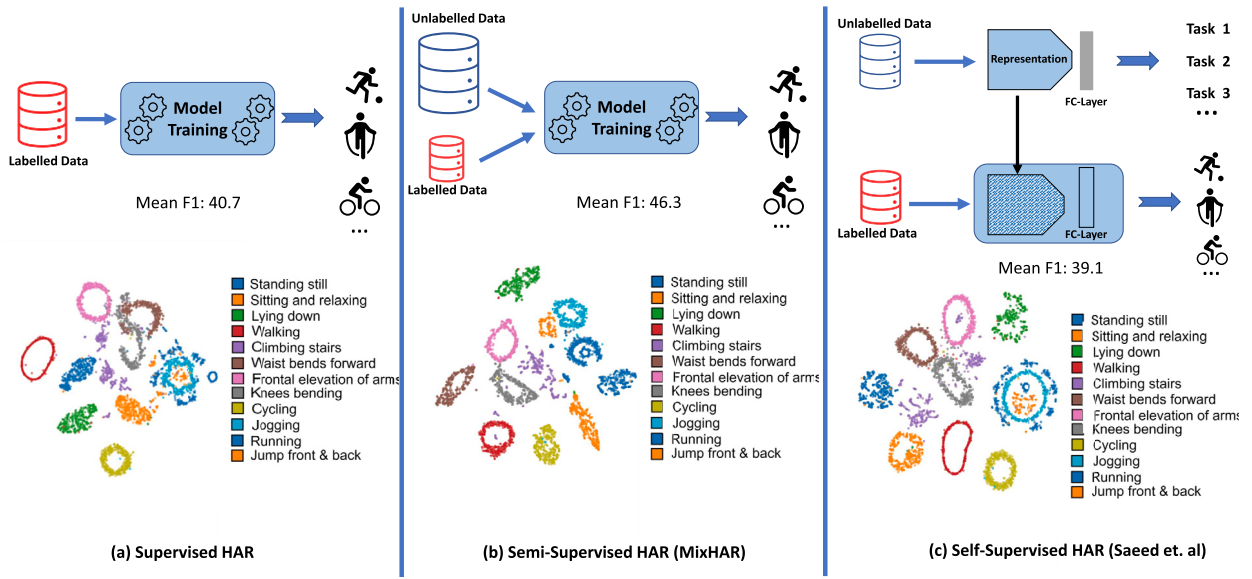

**Fig. 1.** Comparison of the feature diagram of different learning paradigms on deep HAR based on the same Convolution Neural Network, mHealth dataset, and 1% labelled data. The performance of supervised HAR is generally good(a), while the over-fitting still occurred. Self-supervised HAR [5] has pluses and minuses that may not adequately boost performance. As we can see, although it improves the model's performance on waist bends forward and knee bending, the performance is degraded (with a lower mean F1 score) to handle the inter-/intra-activity variability on other activities. Our deep semi-supervised HAR, MixHAR, can clearly reduce the intra-activity distance and enlarge the inter-activity distance with a better mean F1 score by using unlabelled data. Note that in an ideal feature diagram, the same class of activities should aggregate to a single point, and different classes of activities should be dispersed as far as possible.

The alternative is a deep semi-supervised learning approach, which can learn representation from the labelled and unlabelled data simultaneously [11]. With both labelled and unlabelled data and without the requirement of pre-train tasks design, it may provide a more flexible and reliable solution for HAR, as well as the HAR model's performance can be boosted by coordinating the limited yet discriminant activities with extra diverse information (in Fig. 1(b), we show an example of its effectiveness). Although there are some existing semi-supervised HAR works [12,13], there is a clear gap in the state-of-the-art of deep semi-supervised learning HAR and ubiquitous computing HAR. The HAR data, distinct from generic datasets such as data of computer vision tasks, possesses unique spatiotemporal characteristics and temporal dependencies, thus less informative-grained yet such a complex high-level representation requires more data or better design models/methods. One of our goals is to rigorously explore an effective way to take advantage of unlabelled activities data in a deep semi-supervised learning for ubiquitous HAR (i.e., Deep-Semi-HAR).

In this paper, we summarise and define a deep semi-supervised HAR research pipeline. We developed and evaluated five conventional/popular deep semi-supervised techniques on existing HAR works [14]. Most of them focus on regularising the consistency in different perturbed unlabelled data, while they may suffer from non-HAR-specialise perturbations. Also, such consistency regularisation helps to handle the intra-activities variability while ignoring the inter-activities variability.

Motivated by these observations and to better take advantage of wearable-based unlabelled activities, we proposed a deep semi-supervised HAR approach called MixHAR, which is inspired by recent semi-supervised methods like MixMatch [15] and uses unlabelled data to train a robust model by blending labelled and unlabelled activities through linear interpolation, accounting for inter-/intra-activity variability. The transitions between the labelled and unlabelled data incentivise robust network training. Although the linear interpolation method is widely used in various other research tasks [16,17], directly using them with HAR may not be appropriate since the sensor-based human activities hold unique characteristics. Thus, we firstly seek an appropriate way to mix the labelled and unlabelled activities in deep semi-supervised learning. However, we observed an activity-intrusion phenomenon while mixing the activities, i.e., there were occasional conflicts between the mixed activities and the original activities. Hence, eliminating the activity-intrusion problem during the labelled/unlabelled data interpolation is crucial for the completeness of our MixHAR to take advantage of unlabelled data better. Therefore, we propose a mixing calibration mechanism to alleviate the activity-intrusion problem on the feature embedding space. The final MixHAR achieved significant performance and showed the effectiveness of us-ing deep semi-supervised learning techniques to take advantage of unlabelled activities. Our main contributions are summarised as follows.

---

[1] Both [5] and [10] are the methods that aim to take advantage of unlabelled data. However, for a clear and fair comparison, we choose [5].

- We re-produced and evaluated five representative deep semi-supervised learning techniques on wearable-based HAR. Results showed the feasibility of these five Deep-Semi-HAR methods to use unlabelled activities, which paves the way for future deep semi-supervised HAR research.
- On top of our Deep-Semi-HAR setting, we proposed a novel framework called MixHAR that at first generates pseudo labels for unlabelled activities with entropy minimisation. Then, mixes labelled/unlabelled activities by linear interpolation to train the model simultaneously, leading to the better handling of diverse training variability (inter/intra-activities variability).
- We identify an activity-intrusion problem, and we propose a calibration attention mechanism to alleviate the activity-intrusion problem by exploring the correlation on feature space for mixed activities. Our results showed that MixHAR outperformed the five conventional/popular Deep-Semi-HAR.

**2. Background and related work**

This work aims to alleviate the over-fitting of deep HAR and take advantage of unlabelled data and limited labelled data simultaneously in deep semi-supervised learning. This section will cover the algorithmic background to establish the context for both deep HAR and deep semi-supervised learning.

*2.1. Human activity recognition*

Human activity recognition (HAR) can be divided but not limited to two main research directions: vision-based HAR and sensor-based HAR. For vision-based HAR, the data are substantially recorded as video frames, skeleton modality, etc. [18]. Earlier approaches proposed to map the information of activities into 1D hand-crafted features, representing some points of interest that may significantly change in temporal and spatial space [19]. Recent research focused on designing deep learning models to automatically extract the spatial-temporal features, such as transformer [19], LSTM-based models [20] and 3D-CNN-based models [21].

Sensor-based HAR [22], compared with vision-based HAR, can preserve the individual's privacy efficiently, saving computational cost, etc. Using wearable/mobile sensing data for human activity recognition has a long-standing history in the ubiquitous computing community. Early sensor-based HAR approaches tried to utilise the single accelerometer signals to recognise human activities [3]. With the success of feature extraction and variability handling, deep learning has been the mainstream to extract the appropriate features in the human activity recognition system. Guan and Plötz [23] proposed a simple yet powerful deep ensemble framework to combine strong LSTMs with low bias and high variance into a robust learner with variance reduction. Most recently, a discriminative adversarial multi-view network [24] was proposed to model the multi-modal spatial-temporal patterns from the sensory data. This paper proposed a simple but effective Deep-Semi-HAR framework to alleviate the training over-fitting on limited labelled activities by taking advantage of extra unlabelled data.

*2.2. Deep semi-supervised learning*

Deep learning methods provide remarkable improvement for pattern recognition, yet the lack of training data may cause the over-fitting problem. Although extensively collected data may improve the models' generalisation ability, the annotation can be expensive. On the other hand, unlabelled data can be easily acquired, which can be employed for representation learning.

Deep semi-supervised learning paradigm is effective in training models on data that has a few labels but is mostly unlabelled. Deep semi-supervised learning can be used if the data has any of the following three main assumptions [11]: (1) if data points lie in the same cluster in feature space, they should likely be the same class; (2) if data points belong to the same classes or clusters, their outputs from the deep model should be close; and (3) the high-dimensional features of data should roughly lie on a low-dimensional manifold and the classification boundary should not cross the high-density regions. These three assumptions are suitable for most standard classification tasks, which should also be suitable for the sensor-based HAR [25]. Based on these assumptions, it is expected that if a perturbation is applied to the unlabelled data points, the corresponding outputs should still lie close, which is formed as consistency regularisation that most semi-supervised methods rely on [14]. Furthermore, the proxy-label [26] method generates pseudo labels for unlabelled data as additional training examples based on some heuristic. Also, generative models [27] can be used to learn the data distributions from massive unlabelled data and transfer to downstream tasks with limited labelled data [9], which is also known as self-supervised learning or transfer learning.

*Semi-Supervised HAR* using wearable sensors has seen some advancements through semi-supervised learning techniques. These methods leverage both labelled and unlabelled data, offering enhanced performance, especially in situations where labelled data is scarce or costly to obtain. Early works, such as the study by [25], focused on utilizing low-level sensor features from both labelled and unlabelled data. This approach helped in managing noisy data and addressing intra-class variations. Another notable work is by Zeng et al. [28] that involves the use of CNN-Ladder architectures, which allowed for the direct extraction of activity features from sensor data without prior feature engineering. Following this work, Ma et al. [29] proposed a novel forest-style propagation algorithm. This method facilitated the interaction of knowledge between labelled and unlabelled data, enhancing label augmentation for training. Subsequent studies (e.g., [30] introduce the concept of using pseudo-labels, predicted by initially supervised-trained models, in conjunction with labelled data. This trend continued with more recent works [31], employing clustering methods to gather class information for training recognition models.

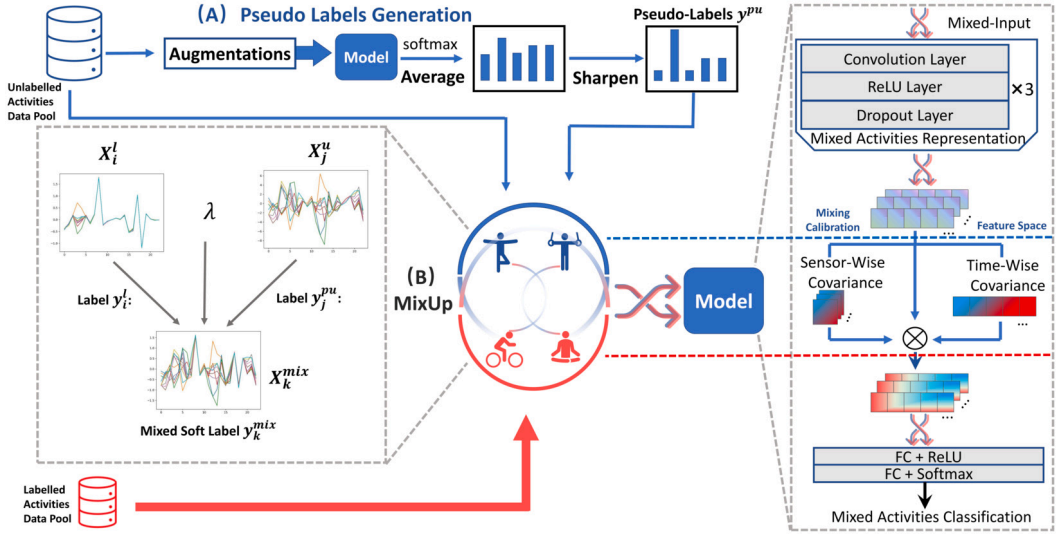

**Fig. 2.** Overview of our proposed method. We first generate the pseudo labels for unlabelled activities and then mix them with labelled activities. The mixed activities are utilised to train the model directly. A mixing calibration mechanism is applied inside the feature space of mixed samples and placed between the representation learning module and the classification module. The $\mathbf{X}_i^\ell$ denote an example of labelled activities with corresponding label $\mathbf{y}_i^\ell$. The $\mathbf{X}_j^u$ denote an example of unlabelled activities and $\mathbf{y}_j^{pu}$ is the corresponding pseudo label. The $\mathbf{y}_k^{mix}$ and $\mathbf{X}_k^{mix}$ denote an example of mixed activities.

The recent research in semi-supervised HAR has shifted towards mining additional useful information from unlabelled data to aid in model training. For instance, [32] introduced a mutual learning framework with a context-aware aggregation module, enabling more reliable sequence information extraction from various sensor data sequences. Zilelioglu et al. [13] advanced the field further by proposing a semi-supervised Generative Adversarial Network, which incorporates temporal information to enrich the feature space from class boundaries. Moreover, the application of semi-supervised HAR has expanded beyond traditional domains. Some studies have extended its use to federated learning [33,34], multimodal systems [35], healthcare-related tasks like stress detection [36], and infrared-based human activity recognition [37].

The evolution of semi-supervised HAR techniques demonstrates a growing sophistication in handling wearable sensor data, and our proposed method builds upon these foundations. We rigorously explore one of the challenges in the field, which is the lack of an effective and equitable benchmarking for leveraging unlabelled activity data, an area that draws inspiration from the broader realm of deep semi-supervised learning (e.g., data split strategies and comparative analysis of methods). Additionally, we introduce a novel pseudo-label-based approach. Our method distinctively uses pseudo labels generated through reliable data mixing techniques to manage the inherent variation in activities. This approach differs from previous works by not only employing pseudo labels but delving deeper into their integration through a mixing module. We complement this with an attention-driven network specifically designed to enhance feature extraction for more accurate activity recognition.

## 3. Methodology

Fig. 2 shows the overview of our methods, which aim to take advantage of substantial unlabelled activities in a semi-supervised fashion. MixHAR first generates pseudo labels for unlabelled activities with entropy minimisation. Then, labelled/unlabelled activities were mixed by linear interpolation to train the model simultaneously. The mixing interpolation suffered from the conflicts between the mixed activities and original activities, which we identified as the activity-intrusion problem. Here, we proposed a calibration attention mechanism by exploring the correlation on feature space for mixed activities to alleviate the activity-intrusion problem, which further enhanced the model's capability of handling the substantial variability from mixed activities.

### 3.1. Deep-semi-HAR pipeline

In supervised HAR, the training data consists of the time series sensor readings of different activities as $\{\mathbf{X}, \mathbf{y}\}$, where $\mathbf{X} \in \mathbb{R}^{H \times T}$ and $\mathbf{y} \in \{1, 2, ..., C\}^T$, $H$ is the number of sensor channels, $T$ denotes the number of temporal samples and $C$ is the total number of classes. Following previous work in HAR [3], the data are segmented into $N$ frames along each sensor channel by a sliding window (with fixed length) and overlapping rate. Then we can obtain the frame-wise time series data $\mathcal{D} = \{(\mathbf{X}_i, y_i)\}_{i=1}^N$, where $\mathbf{X}_i \in \mathbb{R}^{H \times L}$, and the label $y_i \in \{1, 2, ..., C\}$, where $L$ is the length of sliding window. The testing data is processed with the same operation. The deep HAR model with activity representation learning module $\mathcal{F}$ and classification module $\mathcal{S}$ can be end-to-end trained by a loss function (e.g., Cross Entropy Loss) using training data $\mathcal{D}$, and evaluated on unseen testing data.

Then, we define the protocol to split the labelled/unlabelled data for Deep-Semi-HAR (including our proposed MixHAR). The training data $\{\mathbf{X}, \mathbf{y}\}$ can be split into two parts, namely labelled part $\{\mathbf{X}^\ell, \mathbf{y}^\ell\}$ and unlabelled part $\{\mathbf{X}^u, \mathbf{y}^u\}$. A certain percentage (e.g., 1%, 3%, etc.) of the entire training data is used as labelled data, and the rest is utilised as unlabelled data. More specifically, the percentage partition is applied to each class. For instance, each class in the labelled data pool consists of a certain number of activities from the corresponding class in the entire training data. Therefore, the class balance/imbalance distributions in labelled and unlabelled parts are the same as in the original training dataset. To avoid information leaking, labelled/unlabelled data partition is applied *before* the aforementioned data segmentation, which leads to the labelled activity data $\mathcal{D}^\ell = \left\{(\mathbf{X}_j^\ell, y_j^\ell)\right\}_{j=1}^{N^\ell}$ and unlabelled activity data $\mathcal{D}^u = \left\{(\mathbf{X}_k^u)\right\}_{k=1}^{N^u}$. We set $N^u >> N^\ell$ [14] to mimic the practical semi-supervised HAR scenario. In supervised baseline, deep HAR model training only relies on the labelled data $\mathcal{D}^\ell$. Using the labelled data $\mathcal{D}^\ell$, Deep-Semi-HAR also makes use of the unlabelled data $\mathcal{D}^u$, specifically the labelled and unlabelled data are used to train the model *simultaneously* in an end-to-end manner, which aims to alleviate the over-fitting and improve the model's generalisation (one example is shown in Fig. 1).

### 3.2. Labelled and unlabelled activities mixing

Following deep semi-supervised learning, the labelled and unlabelled data are leveraged to train the model simultaneously in our framework. Our approach is motivated by previous work MixUp, which is to mix two samples by linear interpolation, and we extend the MixUp to interpolate the information between existing limited labelled and extra unlabelled activities. In our work, MixUp acts as a data augmentation module and we leverage the mixed new data (activities), which may contain not only discriminant classification information but also inject diverse variability to regularise the model learning to be more robust. A more robust model can be obtained in deep HAR by handling the inter/intra-activity variability. Inter-activity variability occurs when the activities belong to different classes but show similar characteristics (e.g., knee bending and climbing stairs). Intra-activity variability occurs when the same activities are being performed by different subjects or even by the same subject but within the influence of emotional or environmental factors.

When mixing the unlabelled activities with labelled activities together in MixHAR, two cases may lead the deep HAR model to be more robust: (1) if labelled and unlabelled activities are mixed and they belong to different activities, the model's generalisation can be improved by training with injecting diverse inter-activity variability; and (2) if labelled and unlabelled activities are mixed and they belong to same activities but show different patterns, the model's generalisation can be improved by training with injecting diverse intra-activity variability. Furthermore, through the interpolation between labelled and unlabelled data, the network can be incentivised to learn the transitions between each pair-wise sample (that is used to mix). The transitions connect the existing supervised classification information from labelled data and more diverse activity variability from unlabelled data. Rather than using labelled data and unlabelled data separately, our approach helps the deep HAR model to learn the discriminant patterns of different activities while handling large variability simultaneously.

Given a batch (batch size is $B$) of labelled data $\mathcal{B} = \left\{(\mathbf{X}_j^\ell, y_j^\ell)\right\}_{j=1}^{B}$, $\mathcal{B} \in \mathcal{D}^\ell$ and a same sized batch of unlabelled data $\mathcal{U} = \left\{(\mathbf{X}_k^u)\right\}_{k=1}^{B}$, $\mathcal{U} \in \mathcal{D}^u$, we firstly generate the pseudo labels for unlabelled data (Fig. 2(A)). The pseudo labels should be at least more accurate than manually defined classification thresholds and give relatively clear guidance to the model when it is trained with different variability. Motivated by the previous work Unsupervised Data Augmentation [38] that leveraged the multiple augmentations to inject noisy variability and obtained robust pseudo labels, two data augmentation functions $\mathcal{A}_1$ and $\mathcal{A}_2$ are applied on unlabelled data. For each[2] unlabelled data $\mathbf{X}_k^u$, the initial pseudo label is firstly generated with deep model $\mathcal{F}$, which is formed as:

$$\mathbf{y}_k^{pu} = \frac{1}{2}(\text{softmax}(\mathcal{F}(\mathcal{A}^1(\mathbf{X}_k^u))) + \text{softmax}(\mathcal{F}(\mathcal{A}^2(\mathbf{X}_k^u)))), \tag{1}$$

where $\mathbf{y}_k^{pu}$ is a categorical probability vector. The HAR model is expected to predict consistent labels for the same activities with different variability. Hence, we use averages on all predictions rather than any single predictions from the activity sample. There exists a large number of diversities/variabilities in unlabelled activities (50%-99% of training data are unlabelled). Without the real guidance/labels, the model's predictions (i.e., pseudo labels) for the unlabelled activities may contain large uncertainties. These uncertainties may lead to ambiguous classification boundaries [11]. Consequently, the class probabilities (the prediction normalised by softmax function) may be close to a uniform style (as shown in Fig. 2(A) and visualisation in Section 5) with relatively high entropy. Such uniform distribution may not provide the discriminative classification information, while we want the pseudo labels to at least represent different activities with proper discrimination. Motivated by the entropy minimisation in deep semi-supervised learning [39], one way to alleviate the uncertainties in prediction is to enforce model output low entropy predictions on unlabelled data. In this work, we apply a Sharpen operation to the categorical distribution to reduce the uncertainty for each $\mathbf{y}_k^{pu}$, which is formed as:

$$\text{Sharpen}\left(\mathbf{y}_k^{pu}, v\right) = \frac{\left(\mathbf{y}_k^{pu}\right)^{\frac{1}{v}}}{\left\|\left(\mathbf{y}_k^{pu}\right)^{\frac{1}{v}}\right\|_1}, \tag{2}$$

---

[2] Note: for demonstration, we only use one sample as an example.

where the output is a 'one-hot' distribution when $\upsilon \to 0$. After the Sharpen operation, the pseudo labels may contain more discriminative activity classification information. The unlabelled data batch with pseudo labels can be obtained as $\mathcal{U} = \{(\mathbf{X}_k^u, \mathbf{y}_k^{pu})\}_{k=1}^B$. Also, the labels of annotated data are changed to be a 'one-hot' encoding such that $\mathcal{B} = \{(\mathbf{X}_j^\ell, \mathbf{y}_j^\ell)\}_{j=1}^B$. Then unlabelled data with two augmentation functions give us $\mathcal{U}^{\mathcal{A}^1} = \{(\mathcal{A}^1(\mathbf{X}_k^u), \mathbf{y}_k^{pu})\}_{k=1}^B$ and $\mathcal{U}^{\mathcal{A}^2} = \{(\mathcal{A}^2(\mathbf{X}_k^u), \mathbf{y}_k^{pu})\}_{k=1}^B$. A super batch can be built as $\mathcal{X} = \mathcal{B} \cup \mathcal{U}^{\mathcal{A}^1} \cup \mathcal{U}^{\mathcal{A}^2}$, where $\mathcal{X} = \{(\mathbf{X}_k^s, \mathbf{y}_k^s)\}_{k=1}^{3B}$. During the training, given the interpolation (i.e., MixUp) parameter $\lambda \& \sim \text{Beta}(\beta, \beta)$ and $\lambda \& = \max(\lambda, 1-\lambda)$, each paired data point $(\mathbf{X}_k^s, \mathbf{y}_k^s)$ from $\mathcal{X}$ and $(\mathbf{X}'_k^s, \mathbf{y}'_k^s)$ from Shuffled $\mathcal{X}$ are mixed (Fig. 2(B)) to be a single data point as follows:

$$\begin{aligned} \mathbf{X}_k^{mix} &= \lambda \mathbf{X}_k^s + (1-\lambda)\mathbf{X}'_k^s \\ \mathbf{y}_k^{mix} &= \lambda \mathbf{y}_k^s + (1-\lambda)\mathbf{y}'_k^s, \end{aligned} \quad (3)$$

where we can obtain the virtual mixed data batch $\mathcal{X}^{mix} = \{(\mathbf{X}_k^{mix}, \mathbf{y}_k^{mix})\}_{k=1}^{3B}$. The virtual mixed activity data are directly used to train the deep HAR model.

### 3.3. Activity intrusion and mixing calibration

Although leveraging the mixed activities helps the HAR model to obtain better generalisation, not all the mixed activities have a positive impact when conflicts occur between the mixed data and original data, yielding biased HAR model training. We defined this problem as an activity-intrusion problem. Intuitively, the activity-intrusion problem occurs when any virtual-mixed activity is similar to a specific real activity but is assigned a different virtual-mixed soft label distribution. For instance, if a labelled activity is walking and an unlabelled activity is running, the resulting mixed activity may very likely be similar to jogging, but the soft label of the mixed activity is mainly assigned as the probabilities of "walking" and "running." Such an activity-intrusion problem may result in the degradation of the model's discrimination/performance, and affect the learning granularity. Here, motivated by previous work [40] that further feature correlation exploration may alleviate the intrusion problem when a model is trained with MixUped samples, thus we propose a mixing calibration mechanism that learns the feature correlation inside the embedding space, where the correlation is the high order statistics (e.g., covariance) [41] derived from the output features, which also reveals the relationship between the pair-wise mixed activities. Then, the obtained correlation is formed as an attention-guiding model for more discriminant representations. Each virtual mixed activity sample is input to the representation learning module to obtain the activity representation $\mathbf{F}_k$, where $\mathbf{F}_k = \mathcal{F}(\mathbf{X}_k^{mix})$[3] and $\mathbf{F}_k \in \mathbb{R}^{c \times t}$, the $c$ dimensions contain the multi-sensor information and the $t$ dimensions contains the temporal information, and we compute both sensor-wise and time-wise feature covariance matrix, respectively $\mathbf{M}_c$ and $\mathbf{M}_t$, where they can be calculated as:

$$\mathbf{M}_c = \mathbf{F}_k \, \bar{\mathbf{I}}_c \, \mathbf{F}_k^T, \qquad \mathbf{M}_t = \mathbf{F}_k^T \, \bar{\mathbf{I}}_t \, \mathbf{F}_k, \quad (4)$$

where $\bar{\mathbf{I}}_c = \frac{1}{t \times 1}(\mathbf{I}_c - \frac{1}{t \times 1}\mathbf{1}_c)$, $\mathbf{I}_c \in \mathbb{R}^{t \times t}$, $\mathbf{1}_c \in \mathbb{R}^{t \times t}$ and $\bar{\mathbf{I}}_t = \frac{1}{t \times 1}(\mathbf{I}_t - \frac{1}{t \times 1}\mathbf{1}_t)$, $\mathbf{I}_t \in \mathbb{R}^{c \times c}$, $\mathbf{1}_t \in \mathbb{R}^{c \times c}$. The obtained covariance matrix $\mathbf{M}_c \in \mathbb{R}^{c \times c}$ and $\mathbf{M}_t \in \mathbb{R}^{t \times t}$ have clear structural information that each row or column[4] contains statistical dependency of the features along the sensors and times, which also indicates the feature correlation of each pair of activities in a mixed sample. To learn such structural correlation information in an end-to-end manner, we apply a group convolution directly on the obtained covariance matrix by utilising independent convolution filters for each row or column (here, we apply it on each row). The filter size, number of filters and number of groups[5] are all set as the $c$ for $\mathbf{M}_c$ or $t$ for $\mathbf{M}_t$, so $\mathbf{M}_c \in \mathbb{R}^{c \times c}$ and $\mathbf{M}_t \in \mathbb{R}^{t \times t}$ can be derived into $\mathbf{M}_c \in \mathbb{R}^{c \times 1}$ and $\mathbf{M}_t \in \mathbb{R}^{t \times 1}$. Then one standard convolution (filter size is 1 and number of filters are $c$ or $t$) is applied to enhance the correlation learning and the resulting $\mathbf{M}_c \in \mathbb{R}^{c \times 1}$ and $\mathbf{M}_t \in \mathbb{R}^{t \times 1}$ are activated by sigmoid function, which acts as the attention weight vector. Then we reshape the $\mathbf{M}_t \in \mathbb{R}^{t \times 1}$ to be $\mathbf{M}_t \in \mathbb{R}^{1 \times t}$, and the mixing calibration attention can be formed as $\mathbf{M}_c \times \mathbf{M}_t$, and final activity representation can be calculated as $\mathbf{F}_k \otimes (\mathbf{M}_c \times \mathbf{M}_t)$, where the $\otimes$ denotes the multiplication operation. The final representation is input to the classification module $\mathcal{S}$ to make the prediction. Mixing calibration can alleviate the activity-intrusion problem, further enhance/rectify the model's discrimination, and improve the model's capability of handling the substantial injected/mixed variability from virtual-mixed activities.

Finally, in each training iteration, if $\mathbf{X}_k^s \in \mathcal{B}$ or $\mathbf{X}'_k^s \in \mathcal{B}$, we calculate the cross-entropy loss as a lot of information comes from labelled data with smoothly injecting the information from unlabelled data. If $\mathbf{X}_k^s \notin \mathcal{B}$ and $\mathbf{X}'_k^s \notin \mathcal{B}$, we calculate the mean square error (with a loss weight $\gamma$) as HAR models are mainly learned from unlabelled data with pseudo labels and mean square error is less sensitive to outliers in predictions.

---

[3] Note: for demonstration, we only use one sample as an example.
[4] The covariance matrix is diagonally symmetrical.
[5] Parameters of convolution in Pytorch.

## 4. Experiments

### 4.1. Datasets

Our experimental evaluation is conducted on five benchmark datasets, which correspond to diverse yet typical applications in the field of wearable-based HAR, namely *Opportunity* [42], *PAMAP2* [43], *mHealth* [44], and *DSADS* [45]. Also, *mHealth+* dataset further contains the imbalanced NULL-Class Activity (more than 70% of the entire dataset). Other settings in mHealth+ dataset are similar to mHealth dataset. For mHealth dataset, our goal is to show an evaluation of NULL-Class [23] impact between mHealth and mHealth+ datasets.

Following previous works, we set a sliding window of 1 second with 50% overlapping for Opportunity dataset [23], a small window size of 0.8 seconds and 50% overlapping for DSADS dataset [24], and a large window size of 168-time points and 78% overlapping is applied to others [23].

### 4.2. Evaluation protocol

Following the previous works [25], we apply the Leave One Subject Out Cross Validation (LOSO-CV) as an evaluation strategy, where the test data is the activities from the leave one out the subject, and the final results are the average (with the standard deviation) over iterating all the subjects. Also, we use the mean F1 score ($F_m$) as evaluation metrics to measure the performance of different methods, which is calculated as:

$$F_m = \frac{1}{C} \sum_{c=1}^{C} \frac{2\text{TP}_c}{2\text{TP}_c + \text{FP}_c + \text{FN}_c}, \tag{5}$$

where $C$ is the total number of classes, $\text{TP}_c$ is the true positive of each class, $\text{FP}_c$ is the false positive of each class, $\text{FN}_c$ is the false negative of each class.

### 4.3. Deep-semi-HAR baselines

Deep semi-supervised learning has recently achieved state-of-the-art in different research areas [11]. Most of them are based on five conventional/popular works [11,14]. To bridge the gap between recent deep semi-supervised learning and ubiquitous HAR while considering the compatibility, we reproduced and developed the five conventional methods for existing HAR works. These methods solely involved predicting pseudo labels for unlabelled data or conducting regularisation terms using unlabelled data. The comprehensive description/analysis of each method is out of the scope of this paper, and we refer interested readers to [11]. Instead, in this paper, we aim to apply and evaluate these five conventional deep semi-supervised techniques on HAR, and we follow works [11] to conduct the experiments. Most importantly, some of these methods may be sensitive to the training hyper-parameters, which we carefully tuned and listed the values/settings here for the reproducibility of our results.

$\pi$-**Model** [46] is a training framework with two terms of losses. One loss is the standard Cross-Entropy loss based on the limited labelled activities; another one conducts the consistency regularisation between model predictions on unlabelled activities and the model predictions on augmented unlabelled activities. The consistency regularisation in this approach may help the model to be more robust to variability in the same activity. The data augmentation used here is time series scaling; the consistency regularisation loss is based on the mean square error with the weight $\gamma$. Different augmentations may affect the model performance, while the HAR-specific augmentation may be explored in future work.

Virtual Adversarial Training [39] (**VAT**) automatically approximates a perturbation for each unlabelled activity in an adversarial direction. The adversarial direction is the direction in which the label probabilities are most sensitive in input space. Intuitively, this adversarial approximation helps the model learn invariant characteristics from the activities with different variability. In original works, three hyper-parameters may affect the model performance. Motivated by original works [39], in our experiments for VAT, the number of iterations to find the adversarial direction is set to 1, the perturbation size for adversarial direction is set as 10, and the regularisation coefficient that controls the output adversarial value is set as 0.8, and weight for consistency regularisation term on unlabelled data is set as $\gamma$.

Entropy Minimisation [47] is the method that adds a loss term into a standard deep semi-supervised method that encourages the low-entropy prediction from the deep model. We follow the previous work to use the optimal solution that combines it with VAT as **VATENT** [39]. The entropy minimisation loss is directly added to the VAT loss group without weight control.

Rather than directly using the same model on both labelled and unlabelled data, in Mean-Teacher [48] (**MT**), an extra model is derived from the existing model for labelled data by applying the Exponential Moving Average technique, and the consistency regularisation is conducted on the outputs between two models with same unlabelled activities. We set the exponential moving average decay as 0.9 and the weight of consistency regularisation as $\gamma$. Furthermore, Pseudo-Labelling [26] (**PL**) aims to produce the pseudo labels for unlabelled activities using the model itself over the iteration of training. Hence, more diverse training data may improve the model's generalisation.

We also compare our method with the state-of-the-art **MixMatch** work [15], which is most similar to our proposed works. The proposed MixHAR (w/o Cal) is the version that is carefully adapted from the MixMatch for human activity recognition. Hence, we did not additionally list the MixMatch in our comparison. In each epoch, the model is first trained with labelled activities and then predicts the pseudo labels for unlabelled activities. The labelled activities with labels and unlabelled activities with pseudo labels are used all together to optimise the model (with cross-entropy loss) in the next epoch under the standard supervised training scheme. The model architecture used here is the same as the model used both in the supervised baseline and our MixHAR. Following the most deep semi-supervised works [11,14], the weight $\gamma$ follows the ramp-up strategy from zero. They are all trained 150 epochs with a learning rate of $10^{-3}$ and Adam optimiser.

*4.4. Implementation details*

We conducted our experiments by using Python and Pytorch on Ubuntu platform with NVIDIA RTX TITAN GPU. Motivated by recent works [5], there are 3 activity representation encoding blocks and two fully connected layers used as activity classification modules in our HAR model. Each block that encodes the representation consists of a single 1D convolution layer with ReLU activation and dropout techniques (with a rate of 0.3). The number of convolution filters is set as 32, 64, 96 and filter sizes are 5, 5 and 3 for convolution in these 3 blocks. The output from the first fully connected layers consists of 64 units and is activated by the ReLU function. The output from the second fully connected layer consists of the units that reflect the number of classes, which is input to a softmax function to obtain the classification probability.

The mixing calibration is placed before the activity classification module and the details of dynamic calculation are presented in Section 3.3. We followed the standard mini-batch training with setting batch size 32, small batch size is used as the amount of labelled data are small. In each iteration, we sampled two batches of activity frames (with the same batch size), one batch from the labelled data pool and another batch from the unlabelled data pool. Since the total number of labelled data is far less than the total number of unlabelled data, we followed the previous work [11], in which we cyclically sampled the labelled batches until all the unlabelled data were used once.

With Leave-One-Subject-Out Cross Validation (LOSO-CV), the deep network is trained 150 epochs by Adam gradient decent optimiser with learning rate $10^{-3}$. We use the Re-scaling and TimeWarping augmentation for the pseudo-label generation parts, and the $\beta$ is set as 0.8 for data mixing, and $v$ is set as 0.4 for Sharpen operation. Following previous works [11], we set the aforementioned loss weight $\gamma$ in MixHAR and Deep-Semi-HAR baselines by following a ramp-up training strategy.

## 5. Results and discussion

In this section, we provide an in-depth discussion of the performance of our MixHAR.

*5.1. Comparison of different deep-semi-HAR*

Table 1 shows the comparison of different Deep-semi-HAR performances. We summarised the vital observation as follows:

In Table 1, the models in the supervised learning set are baselines and were only trained on labelled data. Compared with the supervised learning baseline models, the five conventional deep semi-supervised techniques helped the models train on labelled/unlabelled data settings, specifically on the dataset with repetitive activities and without NULL-Class activities, such as in DSADS and mHealth datasets. Nevertheless, the performance of these five deep semi-supervised methods was not always satisfactory. This may be because these methods mostly focuses on alleviating the intra-activity variability. When with less labelled data, the primary target may be understanding the current activity; the inter-activity variability may be more crucial for model training. For example, in the DSADS dataset, the users are asked to perform different activities in their own style freely [45], which results in relatively larger intra-activities variability, so these methods obtained obvious effect on the DSADS dataset. In the Opportunity dataset, data collected from only 4 users in a constrained environment may not cause too many intra-activities issues, yet the main challenge may be the discrimination of the different non-repetitive activities. Although the performance of these conventional Deep-Semi-HAR was not always outstanding, results still showed their feasibility of making use of unlabelled activities, and they can be further improved in future work.

Compared with the supervised learning baseline models on all different datasets with different labelled/unlabelled settings, our proposed MixHAR obtained significant improvement in mean F1 score and satisfactory standard deviation reduction, which suggests the advance of our method in taking advantage of unlabelled data in a deep semi-supervised fashion. We can see 0.6%-3.3% $F_m$ improvement on Opportunity dataset, 0.8%-11.1% $F_m$ improvement on PAMAP2 dataset, 5.3%-12.8% $F_m$ improvement on mHealth dataset, 3.8%-10.6% $F_m$ improvement on mHealth+ dataset, and 3.9%-9.7% $F_m$ improvement on DSADS dataset. The improvement on the Opportunity dataset was not large as others, the possible reason may be that the activities in Opportunity are non-repetitive and extremely imbalanced.

Compared with all the Deep-Semi-HAR baselines [46,48,26,39], MixHAR still outperformed them with an obvious gap of $F_m$. Standard deviation indicates the model performance fluctuating on different subjects. Since these five approaches mainly focused on different consistency regularisation for intra-activity variability, they obtained slightly better standard deviation in some cases than ours.

**Effect of mixing calibration.** Table 1 also shows the effect of the proposed mixing calibration. As we can see, the main performance boosting in our MixHAR came from the labelled/unlabelled data mixing (MixHAR w/o Cal in Table 1), which is reasonable as our goal is to take advantage of unlabelled data. With the help of the calibration for the activity-intrusion problem in feature space, we can see MixHAR obtained the satisfactory improvement of 0.1%-5.7% on different datasets under different settings, which suggests the importance of the calibration for labelled and unlabelled data interpolation.

**Table 1**
Comparison of different HAR approaches on different datasets, the percentage denotes the amount of labelled data partitioned from the training data. In bold are the best performance ($F_m$) of MixHAR.

| | | 0.5% | 1% | 3% | 5% | 10% | 30% | 50% |
|---|---|---|---|---|---|---|---|---|
| PAMAP2 | Supervised | 26.7(8.2) | 34.0(8.5) | 39.9(12.7) | 43.8(11.9) | 44.1(13.6) | 74.2(17.7) | 82.5(14.5) |
| | MT-HAR | 17.5(9.3) | 28.0(8.9) | 33.5(13.3) | 42.5(13.6) | 46.2(16.3) | 74.6(16.5) | 82.9(14.6) |
| | Pi-HAR | 27.4(11.1) | 36.0(8.4) | 46.4(11.4) | 48.5(8.9) | 50.4(15.7) | 74.4(14.2) | 80.6(16.7) |
| | PL-HAR | 27.1(7,9) | 27.0(8.5) | 31.2(10.3) | 34.8(12.9) | 38.9(11.8) | 76.4(14.0) | 78.9(16.4) |
| | VAT-HAR | 20.5(7.1) | 25.3(7.9) | 34.7(12.6) | 35.6(13.6) | 47.9(16.2) | 80.3(16.3) | 82.2(15.4) |
| | VATENT-HAR | 18.5(7.8) | 21.4(6.6) | 20.2(14.3) | 24.9(12.9) | 34.7(15.2) | 66.1(15.4) | 70.3(20.1) |
| | MixHAR(w/o Cal) | 28.0(9.3) | 37.0(8.5) | 42.3(10.2) | 49.9(11.1) | 53.4(13.3) | 76.0(12.1) | 82.7(15.4) |
| | **MixHAR** | **28.1(6.7)** | **38.7(6.7)** | **43.4(12.1)** | **48.9(11.2)** | **55.2(14.2)** | **80.7(12.2)** | **83.3(12.8)** |
| mHealth | Supervised | 38.5(8.0) | 40.7(10.6) | 43.0(13.6) | 44.6(8.2) | 42.3(12.3) | 70.4(15.5) | 86.0(9.3) |
| | MT-HAR | 37.8(6.5) | 43.1(7.5) | 44.0(13.0) | 45.6(8.9) | 39.5(10.1) | 69.7(16.8) | 81.3(7.0) |
| | Pi-HAR | 36.7(6.9) | 40.3(8.1) | 45.6(11.6) | 43.1(7.9) | 37.5(13.0) | 71.5(14.3) | 82.8(9.7) |
| | PL-HAR | 39.2(7.0) | 41.8(7.1) | 46.3(11.2) | 46.7(6.8) | 42.0(10.2) | 73.1(13.6) | 83.7(7.4) |
| | VAT-HAR | 38.8(6.5) | 40.3(8.0) | 45.9(10.1) | 44.0(7.0) | 39.1(11.4) | 72.8(13.5) | 80.1(9.9) |
| | VATENT-HAR | 40.0(7.3) | 39.4(7.4) | 43.0(10.5) | 46.6(7.6) | 46.6(6.9) | 76.0(11.3) | 83.2(9.9) |
| | MixHAR(w/o Cal) | 40.1(7.8) | 44.8(10.3) | 51.4(12.1) | 53.2(7.9) | 51.5(9.8) | 82.3(11.5) | 91.4(5.1) |
| | **MixHAR** | **43.8(7.6)** | **46.3(10.3)** | **52.5(11.8)** | **53.8(7.5)** | **55.1(10.1)** | **82.3(10.4)** | **92.3(4.3)** |
| DSADS | Supervised | 41.0(4.9) | 50.6(20.4) | 59.5(8.6) | 61.0(7.4) | 62.8(5.8) | 84.5(7.0) | 86.7(6.3) |
| | MT-HAR | 48.2(2.3) | 56.5(4.8) | 59.4(6.8) | 62.9(7.1) | 61.8(5.6) | 83.9(6.7) | 86.0(7.3) |
| | Pi-HAR | 49.2(4.1) | 57.3(6.5) | 61.4(7.8) | 60.2(7.3) | 63.0(8.0) | 82.2(8.8) | 85.8(9.2) |
| | PL-HAR | 47.1(3.0) | 58.0(5.8) | 63.6(9.3) | 64.4(4.4) | 65.3(6.9) | 81.6(10.1) | 86.1(7.3) |
| | VAT-HAR | 49.1(2.2) | 56.3(7.0) | 58.9(7.3) | 64.2(3.4) | 64.4(8.4) | 81.6(8.1) | 84.9(7.6) |
| | VATENT-HAR | 46.8(3.1) | 56.2(5.7) | 58.1(7.6) | 60.0(7.6) | 61.0(7.2) | 87.3(6.6) | 87.1(8.0) |
| | MixHAR(w/o Cal) | 44.5(2.1) | 58.7(6.6) | 64.6(6.6) | 63.1(6.9) | 66.1(7.2) | 85.8(5.2) | 89.1(4.3) |
| | **MixHAR** | **50.3(4.6)** | **59.5(6.3)** | **66.6(8.4)** | **64.9(5.3)** | **67.8(6.5)** | **89.9(6.9)** | **90.1(7.2)** |
| mHealth+ | Supervised | 13.2(4.1) | 19.6(6.2) | 25.0(8.5) | 26.5(9.6) | 25.8(9.5) | 44.7(14.9) | 52.5(11.0) |
| | MT-HAR | 14.2(3.4) | 24.1(5.0) | 25.2(7.0) | 23.9(7.1) | 20.1(8.5) | 44.8(13.7) | 54.2(11.4) |
| | Pi-HAR | 14.2(2.4) | 24.9(6.6) | 23.8(6.6) | 23.4(5.5) | 20.0(8.4) | 42.3(12.6) | 51.5(13.0) |
| | PL-HAR | 13.7(3.0) | 24.5(6.7) | 24.0(6.8) | 23.5(5.1) | 22.0(9.2) | 42.1(13.7) | 51.5(14.2) |
| | VAT-HAR | 13.2(2.9) | 24.8(7.0) | 24.2(6.0) | 24.6(8.4) | 20.6(7.9) | 40.4(13.2) | 51.4(11.7) |
| | VATENT-HAR | 14.8(4.5) | 25.5(4.3) | 28.0(7.5) | 25.4(5.9) | 28.5(5.1) | 53.5(10.9) | 53.4(8.0) |
| | MixHAR(w/o Cal) | 15.5(4.1) | 28.3(6.2) | 32.9(5.8) | 28.2(7.1) | 30.0(9.2) | 51.8(15.1) | 54.3(12.9) |
| | **MixHAR** | **17.0(3.8)** | **29.2(6.0)** | **33.9(7.6)** | **34.5(6.8)** | **36.3(7.0)** | **55.3(10.6)** | **56.4(7.1)** |
| Opportunity | Supervised | 18.2(3.5) | 23.0(4.0) | 28.2(3.3) | 30.8(2.7) | 32.8(6.0) | 40.3(2.7) | 41.2(3.4) |
| | MT-HAR | 19.1(2.9) | 23.0(4.0) | 28.2(2.6) | 30.5(4.3) | 32.3(5.1) | 38.0(2.0) | 41.0(2.1) |
| | Pi-HAR | 16.1(1.6) | 24.4(3.1) | 27.4(2.8) | 30.1(2.7) | 31.0(4.8) | 39.5(2.6) | 40.0(2.2) |
| | PL-HAR | 15.0(2.2) | 24.0(3.1) | 27.3(3.0) | 30.8(4.2) | 32.9(5.7) | 40.2(3.6) | 41.5(4.7) |
| | VAT-HAR | 17.0(2.3) | 24.5(3.4) | 27.1(2.7) | 29.7(4.4) | 30.9(5.5) | 39.0(2.7) | 40.5(3.0) |
| | VATENT-HAR | 10.3(1.4) | 17.2(3.4) | 26.9(5.6) | 30.4(2.6) | 32.5(5.6) | 39.1(0.8) | 39.4(2.8) |
| | MixHAR(w/o Cal) | 19.3(3.0) | 23.3(3.5) | 29.4(2.8) | 31.8(3.3) | 34.3(4.3) | 40.8(3.8) | 41.3(4.1) |
| | **MixHAR** | **20.1(3.5)** | **25.4(2.5)** | **31.5(1.7)** | **32.0(3.1)** | **35.5(3.9)** | **40.9(3.0)** | **43.4(3.9)** |

**Effect of NULL-Class.** In the real-world scenario of ubiquitous HAR, most collected activities are irrelevant to the HAR system (NULL-Class), and there are only a few activities of interest [3]. There are two factors that the NULL-Class may affect recognition performance (e.g., on Opportunity and mHealth+ datasets). On the one hand, it is extremely imbalanced (e.g., more than 70% of the entire dataset), causing biased learning of the deep HAR model.

Moreover, when the number of labelled data is very small, it is possible that most of the labelled data is NULL-Class activities, yet the activities of interest are the real targets and may very likely be under-represented/ignored during the training, which resulted in low $F_m$. On the other hand, the NULL-Class has similar patterns as activities of interest (but is unrelated to application) [3], which may lead to ambiguity during the model training. The opportunity dataset contains lots of NULL-Class activities, and we can see that the overall mean F1 score of different methods on it was low, while the effect of our method (MixHAR) is obvious. The activities of interest in mHealth are mostly balanced; we used mHealth and its version with NULL-Class (mHealth+) to reveal the effect of NULL-Class. In Table 1, the mean F1 score of different methods on mHealth+ (with NULL-Class) was largely degraded compared with the corresponding setting on the mHealth (without NULL-Class). However, our MixHAR still obtained more impressive performance than the supervised baseline and other Deep-Semi-HAR.

### 5.2. Feature embedding overview

Fig. 3 visualises the learned feature embedding diagram of different Deep-Semi-HAR methods (including proposed MixHAR) based on the mHealth dataset.

**Model's Generalisation.** A robust deep HAR model should be able to handle the inter/intra-activity variability, as reflected in the feature diagram; the same classes of activities should be clustered together as tight as possible and different classes of activities should be separated apart distinctly. The activities in the mHealth dataset are repetitive (e.g., walking, running), as shown in Fig. 3, and more than half of the categories seem easy to discriminate by the supervised baseline. There are obvious cons and pros of $\pi$-Model, PL and MT approaches. They all performed better on recognising the cycling by clear intra-activity distance reduced. However, the injected variability from unlabelled data may be too noisy to handle and misguide the representation learning. VAT and VATENT performed better than the supervised baseline by better handling the intra-activity variability on walking, cycling and climbing stairs. Most importantly, the robustness of the model trained by MixHAR was considerably improved by enlarging the inter-activity distance and reducing the intra-activity distance on all different activities, which suggests the effectiveness of MixHAR in taking advantage of unlabelled data and improving the model's generalisation.

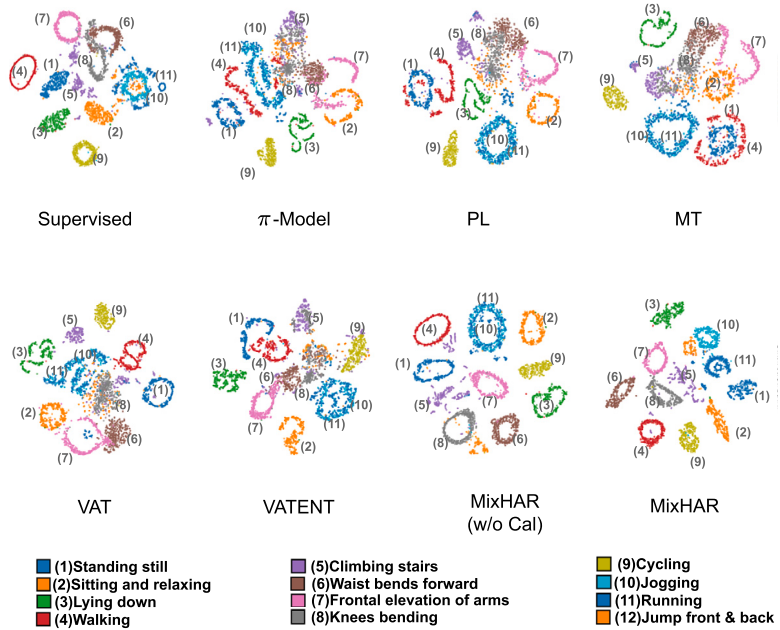

**Fig. 3.** Feature diagram on mHealth dataset based on 1% labelled data setting.

**Effect of Mixing Calibration.** Mixing labelled/unlabelled data injects a large variability for robust model training while keeping discriminative classification information. Since the labelled data are limited and the mixing of substantial unlabelled data may challenge the model's capability of handling diverse information. Specifically, when activity-intrusion happens, the conflicts between the mixed activities and original activities may inhibit the model's discrimination on different activities, which is still weak on different activities with similar characteristics such as knees bending and jumping front & back, running and jogging (MixHAR w/o Cal in Fig. 3). Our proposed mixing calibration aims to learn the correlation of mixed samples on feature space and enhance the model's discrimination on injected variability. With further mixing calibration (MixHAR in Fig. 3), it was very clear that all the different activities (inter-activity) were discriminated obviously by MixHAR with intra-activity distance reduced and inter-activity distance enlarged.

### 5.3. Improvement of minority-activity-classes

Fig. 4 shows the benefits of using MixHAR for each class in the practical case that the data are imbalanced (typically with more than 70% NULL-Class). For both Opportunity and mHealth+ data, using unlabelled data by MixHAR (red/purple in Fig. 4) obtained better generalisation on most minority classes while keeping the performance on majority classes. Although the model's generalisation was improved for most activities by MixHAR for the Opportunity dataset, the model's performance still mainly came from the imbalanced NULL-Class activities. This is because in the Opportunity dataset, most of the NULL-Class activities are walking or other simple transitional activities [42], and other activities of interests are relatively complex (i.e., non-repetitive) [42]. For the mHealth+ dataset, we can observe the obvious improvement of MixHAR. It is interesting that when the labelled data are limited, leveraging extra unlabelled data may still not be effective for activities that are relatively simple, like standing still. These characteristics may widely exist in many other activities, which may be misrecognised.

We also compared our MixHAR with the standard random oversampling (orange in Fig. 4) and undersampling (green in Fig. 4) method. As we can see, oversampling and undersampling are feasible to improve the minority class; however, our method still outperformed them in nearly all the classes of activities. Moreover, oversampling/undersampling methods largely degraded the model performance on the head class (NULL-Class), while our method also improved its performance on it. This result suggests that our deep semi-supervised method, MixHAR, may be able to handle the largely imbalanced activities that can be a valuable research direction in real-world ubiquitous HAR.

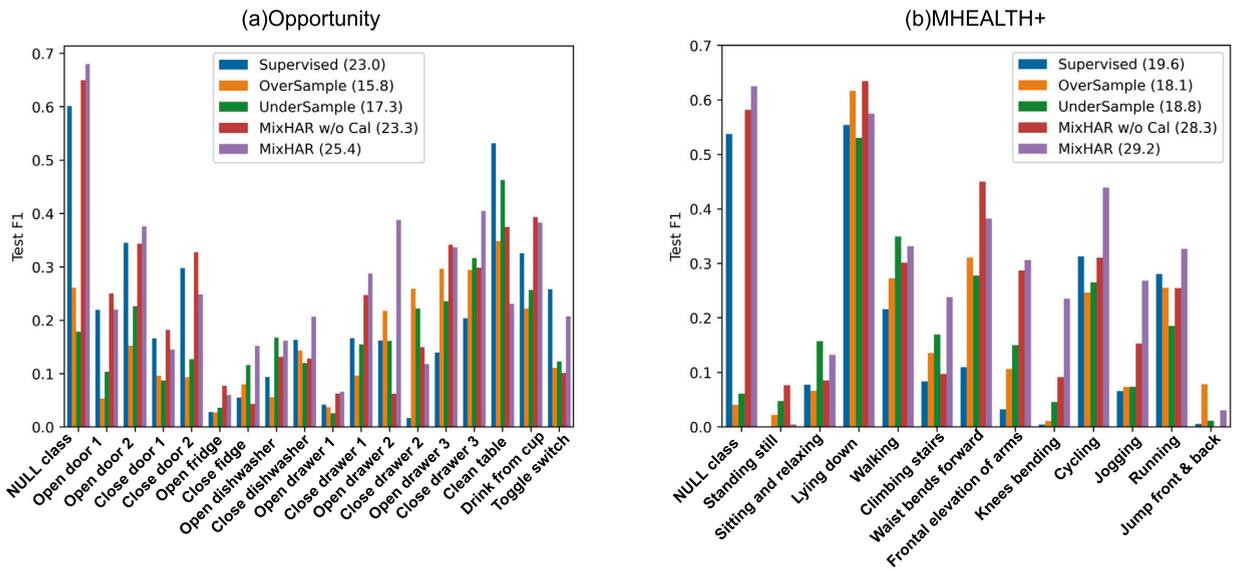

**Fig. 4.** Class-wise recognition results on the data imbalanced dataset Opportunity and mHealth+ based on 1% labelled dataset. The mean F1 score is placed in brackets.

**Table 2**
Comparison of the approaches that take advantage of unlabelled data with 1% labelled data, which is based on our deep semi-supervised settings/pipelines.

| methods | Opportunity | PAMAP2 | mHealth | mHealth+ | DSADS |
| --- | --- | --- | --- | --- | --- |
| AAE-HAR [49] | 10.6(**1.1**) | 16.3(**5.3**) | 26.9(5.8) | 13.1(**0.7**) | 18.4(**2.3**) |
| MTL-Self-HAR [5] | 13.5(1.2) | 18.5(6.57) | 39.1(**5.2**) | 17.9(4.9) | 20.1(3.1) |
| CL-HAR [28] | 11.4(1.5) | 17.3(6.1) | 42.9(10.8) | 24.4(4.7) | 39.4(7.6) |
| Mask-Self-HAR [9] | 15.9(1.76) | 32.8(9.7) | 41.3(8.6) | 25.5(6.8) | 51.4(6.1) |
| Selfhar [10] | 20.9(4.8) | 40.3(12.4) | 44.2(7.9) | 24.6(5.4) | 53.3(1.4) |
| **MixHAR** | **25.4**(2.5) | 38.7(6.7) | **46.3**(10.3) | **29.2**(6.0) | **59.5**(6.3) |

## 5.4. Taking advantage of unlabelled data

In recent years, using unlabelled data has attracted growing research attention in deep HAR [5,9,11]. Specifically, in addition to semi-supervised HAR, self-supervised learning has also obtained rapidly growing attention. To evaluate the advance of our MixHAR, here we also compared MixHAR against the recent state-of-the-art methods taking advantage of unlabelled data, which includes but is not limited to semi-supervised HAR (i.e., self-supervised HAR). We developed the five recent existing works on five wearable-based HAR datasets.

Adversarial Auto-Encoder (AAE-HAR) focuses on learning the activity representations by an encoder-decoder network with two discriminators handling the style information. A multi-task self-supervised learning framework was proposed to pre-train an activity representation learner by discriminating the various transformations that applied on input signals [5], and the representation learner was used on downstream activity classification task (MTL-HAR). Motivated by ladder net [50], Zeng et al. also proposed a CNN encoder-decoder network with noisy injecting and cleaning to exploit unlabelled data [28]. Haresamudram et al. proposed a self-supervised method to pre-train an activity representation learner based on reconstructing unlabelled data in a mask-then-reconstruct manner [9], and then the trained learner was utilised on downstream activity classification task (Mask-Self-HAR). Most recently, Tang et al. proposed a SelfHAR framework based on the existing self-supervised technologies and semi-supervised training to leverage the unlabelled human activities [10].

As shown in Table 2, we can see that our MixHAR outperformed all these methods on the mean F1 score with a slightly higher standard deviation. In AAE, it is hard to judge what information exactly the discriminators learned, it seems that when the number of labelled data is extremely small (e.g., 1% of entire training data) the discriminator may not be fitted and negatively affect the final recognition performance. For the CL-HAR net, the encoder-decoder based on the noisy signals may lose the information of original signals, and the reconstruction process is always biased. Analogously, the MTL-HAR, which built the self-supervised framework based on general data augmentation, may suffer from designing the HAR-irrelevant pre-training tasks. As for Mask-Self-HAR, reconstructing the original signals helped the model to learn the relatively accurate patterns of different activities, which obtained satisfactory performance. Also, the most recent work, i.e., Selfhar, obtained clear improvement compared with previous methods. However, the weak HAR-specialised information during the pre-training may still result in less robust features and not strong enough performance on relatively complex data for these self-supervised based methods. Moreover, when with NULL-Class activities, the activities of interest may be largely under-represented during the pre-training stage. Hence, the performance of self-supervised-based methods is still lower than ours when with non-repetitive activities (e.g., Opportunity dataset) or NULL-Class activities, which suggests the more robustness and reliability of our approach.

## 6. Conclusion

Based on the recent success of semi-supervised learning techniques in deep learning, we developed and evaluated five conventional/popular deep semi-supervised learning approaches for ubiquitous human activity recognition (HAR): Deep-Semi-HAR. Most importantly, we proposed a novel deep semi-supervised approach named MixHAR to take advantage of unlabelled activities. MixHAR leverages linear interpolation to mix the labelled and unlabelled data simultaneously with corresponding mixing calibration, which not only ensures the model's discrimination of different activities but also improves the model's capability to handle diverse variability. Compared with conventional Deep-Semi-HAR methods and recent works that use unlabelled data, our results show that MixHAR is more effective and powerful for sensor-based human activity recognition.

**CRediT authorship contribution statement**

**Haoran Duan:** Writing – original draft, Visualization, Methodology, Formal analysis, Conceptualization. **Shidong Wang:** Validation, Methodology, Formal analysis, Conceptualization. **Varun Ojha:** Writing – review & editing, Visualization, Supervision, Methodology, Conceptualization. **Shizheng Wang:** Visualization, Validation, Resources, Project administration, Data curation. **Yawen Huang:** Writing – review & editing, Visualization, Validation, Resources, Conceptualization. **Yang Long:** Validation, Supervision, Methodology, Formal analysis. **Rajiv Ranjan:** Writing – review & editing, Validation, Resources, Formal analysis, Conceptualization. **Yefeng Zheng:** Writing – review & editing, Visualization, Validation, Resources, Methodology, Formal analysis.

**Declaration of competing interest**

The authors declare that they have no known competing financial interests or personal relationships that could have appeared to influence the work reported in this paper.

**Data availability**

Data will be made available on request.